\documentclass[conference]{IEEEtran}
\IEEEoverridecommandlockouts
\usepackage{cite}
\usepackage{amsmath,amssymb,amsfonts}
\usepackage{algorithm,algpseudocode}
\usepackage{graphicx}
\usepackage{textcomp}
\usepackage{xcolor}
\usepackage{array}
\usepackage{multicol,multirow} 
\usepackage{booktabs}
\usepackage{caption}
\usepackage{tikz,tkz-euclide}
\usepackage{footmisc}
\usepackage{subfigure}
\usepackage{setspace}
\usepackage{url,balance}
\usepackage{ulem}
\usetikzlibrary{quotes,angles}
\usetikzlibrary{calc}
\usetikzlibrary{decorations.pathreplacing}

\def\Est{\mathrm{Estimate}}
\def\Eval{\mathrm{Evaluate}}
\def\TPKL{\mathrm{TPKL}}

\def\emph#1{\textit{#1}}

\def\BibTeX{{\rm B\kern-.05em{\sc i\kern-.025em b}\kern-.08em
    T\kern-.1667em\lower.7ex\hbox{E}\kern-.125emX}}
\begin{document}

\title{Online Game Level Generation from Music
}

\author{
  \IEEEauthorblockN{Ziqi Wang\IEEEauthorrefmark{1}\IEEEauthorrefmark{2}, Jialin Liu\IEEEauthorrefmark{2}\IEEEauthorrefmark{1}}
  \IEEEauthorblockA{
    \IEEEauthorrefmark{1} \textit{Research Institute of Trustworthy Autonomous System}\\
    \textit{Southern University of Science and Technology (SUSTech)}}
    \IEEEauthorblockA{\IEEEauthorrefmark{2}\textit{Guangdong Provincial Key Laboratory of Brain-inspired Intelligent Computation}\\
    \textit{Department of Computer Science and Engineering} \\
    \textit{Southern University of Science and Technology (SUSTech)}\\ Shenzhen, China
  }
}


\maketitle

\begin{abstract}
Game consists of multiple types of content, while the harmony of different content types play an essential role in game design. However, most works on procedural content generation consider only one type of content at a time. In this paper, we propose and formulate online level generation from music, in a way of matching a level feature to a music feature in real-time, while adapting to players' play speed. 
A generic framework named online player-adaptive procedural content generation via reinforcement learning, OPARL for short, is built upon the experience-driven reinforcement learning and controllable reinforcement learning, to enable online level generation from music. Furthermore, a novel control policy based on local search and k-nearest neighbours is proposed and integrated into OPARL to control the level generator considering the play data collected online. Results of simulation-based experiments show that our implementation of OPARL is competent to generate playable levels with difficulty degree matched to the ``energy'' dynamic of music for different artificial players in an online fashion. 

\end{abstract}



\section{Introduction}
Human perception is multi-modal. Digital games, as an emerging creation field, lies in the intersection of multiple types of content that meet different aspects of human perception, has the ability of expressing stories, emotions or aesthetics, and satisfying human's natural entertainment demand~\cite{liapis2018orchestrating,yannakakis2018artificial}. A successful game should guarantee the harmony of different types of content. Procedural content generation (PCG) \cite{shaker2016procedural,togelius2013procedural,summerville2018procedural,liu2021deep}, aiming at the automated or mixed-initiative creation of game contents, such as levels, maps, musics, rules and narratives, has shown its potential to reduce game development costs, augment the creativity of individual human creators, and provide personalised game contents \cite{yannakakis2018artificial,liu2021deep}.
There have been extensive researches in generating a single type of game content~\cite{yannakakis2018artificial,liu2021deep}, while only a few works considered generating one type of content driven by another~\cite{plans2012experience,alamgir2013condition,engels2015automatic}. Procedurally generating a complete game of several content types is an emerging topic~\cite{liu2021deep,hoover2015audioinspace}. For instance, \textit{AudioInSpace}~\cite{hoover2015audioinspace} generated game rules, visuals and audio in a mix-initiative way. The aforementioned works~\cite{plans2012experience,alamgir2013condition,hoover2015audioinspace,engels2015automatic,liapis2018orchestrating} mainly focused on puzzle games, shooting games and rhythm games. 


\begin{figure}[t]
    \centering
    \includegraphics[width=0.95\linewidth]{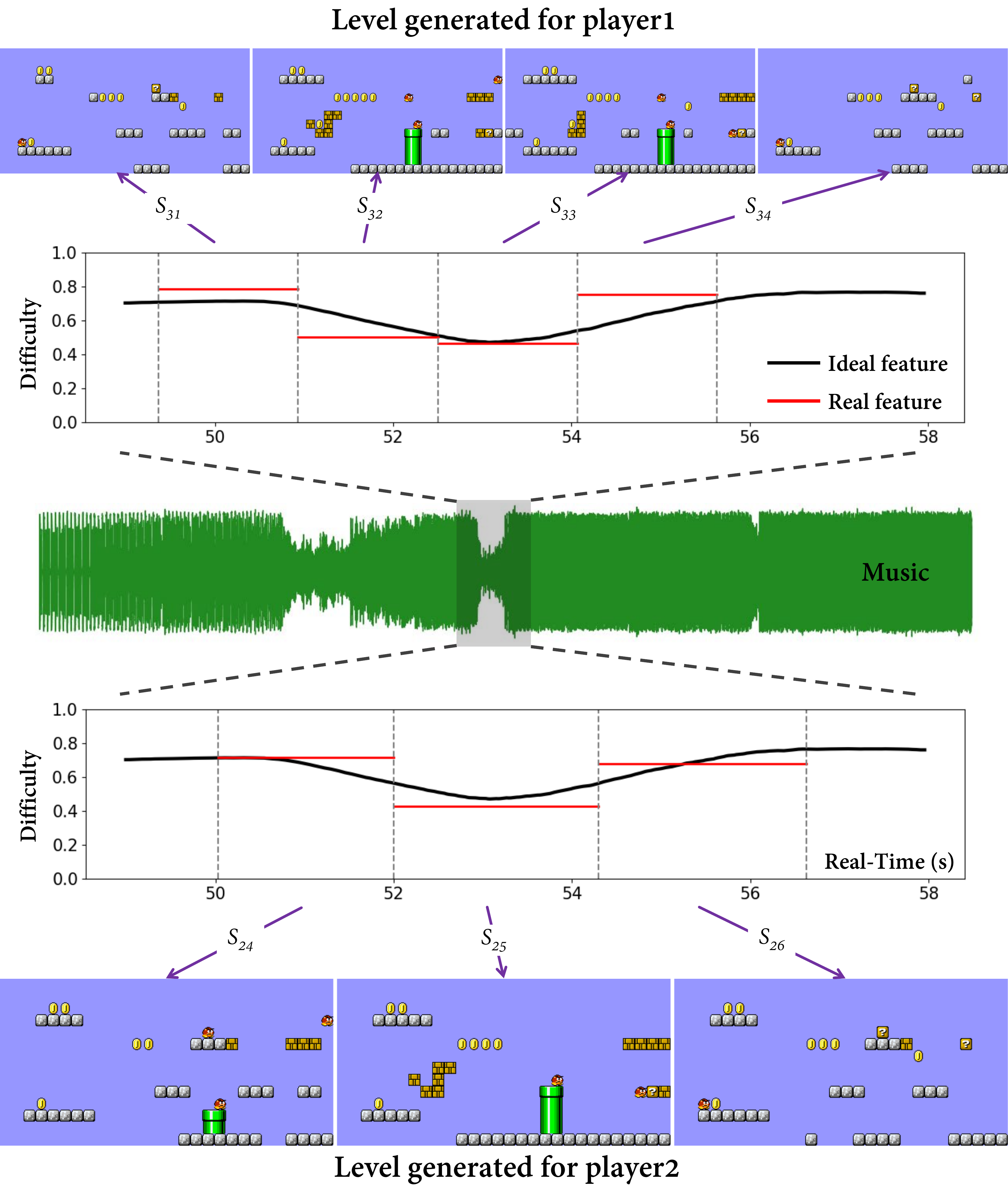}
    \caption{\label{fig:illustration}Illustration of online level generation from music.
    Widths of interval between grey dashed lines correspond to the time consumed to play through a segment.
    This figure is plotted based on the data of simulation-based experiments of this paper (detailed in Section \ref{sec:exp}). The segment-wise difficulty degrees are closed to the ideal difficulty curve in black colour, which is converted from the input music's energy. Moreover, the level generation is adapted to the specific player's play speed in real-time.}
\end{figure}

In this work, we focus on platformer games, in which levels and music together affect the player experience. We investigate into level generation from music, ideally to be achieved in an online fashion for some games, assuming that the play experience should be consistent with the current music. For example, if the music in a platformer game is intense or nervous, the appeared level segments should be more difficult to match the atmosphere created by music. Some commercial games, such as \textit{Dance Dance Revolution} (Konami, 1998), \textit{Guitar Hero} (RedOctane, 2005) and \textit{Muse Dash} (PeroPeroGames, 2018), directly force the gameplay to be consistent with music by designing specific game rules. However, for most games of other genres, it's hard to generate levels from music because the rhythm of playing a level depends on the player's play speed and thus not fixed, which will result in the failure of matching a level to music in an offline or player-unaware way. Fig. \ref{fig:illustration} illustrates why the play speed affects the online generation of level from music.
The level segments on top and bottom are played by different artificial players at the same time window but have different length as the players played the levels in different speed. Moreover, the generation system should determine the difficulty degree of the next level segment in real-time to make sure that the segment will be played at the best-matched time slice. Therefore, online and player-adaptive level generation from music is desired. 

In this paper, a framework for online level generation from music named online player-adaptive procedural content generation via reinforcement learning, OPARL for short, is proposed. OPARL follows a controller-generator architecture. 
The generator takes the current level segment and a music feature as input and outputs a new level segment with feature matched to the input at each iteration. When OPARL generates a level segment from music, the generator is controlled by a novel policy named local search with k-nearest neighbours based estimation (LS-KNN) which takes historical play data into account to produce the targeted feature value that minimises an inner error respect to an ideal feature sequence derived from the given music's temporal feature sequence (detailed in section \ref{sec:mdfc}). The controllable generator is extended from the experience-driven procedural content generation via reinforcement learning (EDRL) \cite{shu2021experience} framework by leveraging controllable reinforcement learning \cite{earle2021learning,khalifa2020pcgrl}.
The proposed OPARL framework is implemented and verified on the benchmark game for level generation, \textit{Super Mario Bros.} (SMB). Experimental results show that the resulted system can generate playable SMB levels that are consistent with the given music\footnote{\label{a}Code, experimental data and demo of this paper are available on GitHub: \url{https://github.com/PneuC/OPARL}.}. 

Our framework requires no expert knowledge expect for the content representation, a few training examples for training a generative adversarial network (GAN) \cite{Goodfellow2014generative,volz2018evolving} and the CNet-assisted repairer \cite{shu2020novel} for determining and repairing broken pipes. Although our proposed approach is verified in generating SMB levels from music, applying it to other platformer games, such as \textit{Megaman} (CAPCOM, 1987), \textit{Electronic Super Joy} (Michael Todd, 2014) and \textit{Celeste} (Matt Makes Games Inc., 2018), is straightforward.

The remainder of this paper is organised as follows. Section \ref{sec:background} presents some related work. Section \ref{sec:mdfc} formulates the problem of online level generation from music and addresses its challenges. Then, our OPARL framework and its technical details are presented in Section \ref{sec:framework}. In Section \ref{sec:exp}, the effectiveness and robustness of OPARL to different players and music are verified through its implementation for SMB and simulation-based experiments. Section \ref{sec:conclusion} concludes and discusses some future directions.

\section{Background}\label{sec:background}
This section discusses related work that involves multiple types of content and online player-adaptive approach. 


\subsection{PCG that Involves Multiple Content Types}

Liapis \textit{et al.} proposed the concept of \textit{orchestrating game generation}, which aims at generating different types of game content jointly and harmoniously \cite{liapis2018orchestrating}. One representative orchestrating game generation system, \textit{AudioInSpace} \cite{hoover2015audioinspace}, generates game rules, visuals and audio in a mix-initiative way. 
Plans and Morelli proposed an experience-driven generator to generate music that reacts to the ``excitement'' of the game using search-based algorithms \cite{plans2012experience}. Naushad and Muhammad introduced a conditional music generation framework to enable adaptive music generation for games \cite{alamgir2013condition}. Engels \textit{et al.} developed an hierarchical Markov model-based music generation system to produce music pieces in real-time \cite{engels2015automatic}.

The aforementioned works mainly focus on generating \textit{cosmetic} content from \textit{functional} content \cite{summerville2018procedural}, while there are also research works that explored the reversed way. For instance, some works concentrated on learning to generate rhythms game charts from music via supervised learning \cite{donahue2017dance,halina2021taikonation}. Jordan \textit{et al.} introduced a mobile game named \textit{BeatTheBeat} which applied self-organising maps method to create game levels that match some music features \cite{jordan2012beatthebeat}. In the work of \cite{karavolos2015mixed}, a mixed-initiative PCG system is presented by Karavolos \textit{et al.} to generate game levels from mission or space provided by human designers. Atmaja \textit{et al.} proposed a top-down PCG framework to ``translate'' platformer games from storyline \cite{atmaja2018generating}.
To our best knowledge, no work has ever generated platformer game levels from music in real-time.

\subsection{Online Player-Adaptive Level Generation}

A popular research topic related to online player-adaptive level generation is dynamic difficult adjustment (DDA), aiming at adjusting levels' difficulty degree considering the abilities or skills of players for desired player experience or aesthetic goal \cite{hunicke2005case}. 
Shi and Chen proposed a DDA policy based on Thompson sampling, and embedded it into an online level generation framework \cite{shi2017learning}. Stammer \textit{et al.} applied a conditional player experience model considering different play styles to generate player-adaptive \textit{Spelunky} levels with DDA \cite{stammer2015player}.

Some works built on experience-driven procedural content generation \cite{yannakakis2011experience}. 
For instance, Shaker \textit{et al.} applied player modelling and grammar evolution to generate online levels that optimise player experience \cite{shaker2012evolving}. Blom \textit{et al.} generated online personalised SMB levels with facial expression recognition \cite{blom2014towards}. Shu \textit{et al.} introduced the EDRL framework to enable real-time level generation with experience-driven reward functions as content quality measurements \cite{shu2021experience}.

\section{Music-driven Online Level Feature Control}\label{sec:mdfc}

To generate level segments that are consistent to a given piece of music in real-time, we consider ensuring the consistency or harmony by matching a feature of level segments to a feature of the given piece of music, referred to as music-driven online level feature control in this paper.
The problem of finding online the optimal value of level feature that matches the given music is formulated in Section \ref{sec:formulation}). Then, the challenges of achieving online music-driven level feature control are discussed in Section \ref{sec:challenges}.

\subsection{Problem Formulation}\label{sec:formulation}

Given a piece of music, an ideal feature sequence of a level can be derived in some fine-grained time unit, denoted as $\mathcal F^* = (f_1^*, \cdots, f_t^*, \cdots)$. Let $\hat f_i$ and $\delta_i$ denote the target feature produced by the controller at the $i$th iteration and the duration of playing the $i$th segment, $S_i$, respectively. The objective is to minimise the error defined as follows:
\begin{equation}
    \label{eq:errin}
    \varepsilon_{inner} = \frac{1}{T(f_{max} - f_{min})} \sum_{i=1}^N \left( \sum_{t=b_i}^{b_i+\delta_i} |f_t^* - \hat f_i|\right),
\end{equation}
where $N$ is the number of segments and $b_i =  \sum_{k=0}^{i-1} \delta_k$ indicates the starting time of playing the level segment $S_i$ with $\delta_0 = 0$. $f_{max}$ and $f_{min}$ are the upper bound and lower bound of the level feature, respectively. $T=\sum_{i=1}^N\delta_i$ is the total time spent to play through the whole level. The decision space of online music-driven level feature control is $[f_{min}, f_{max}]$.

\begin{figure}[htbp]
    \centering
    \includegraphics[width=\linewidth]{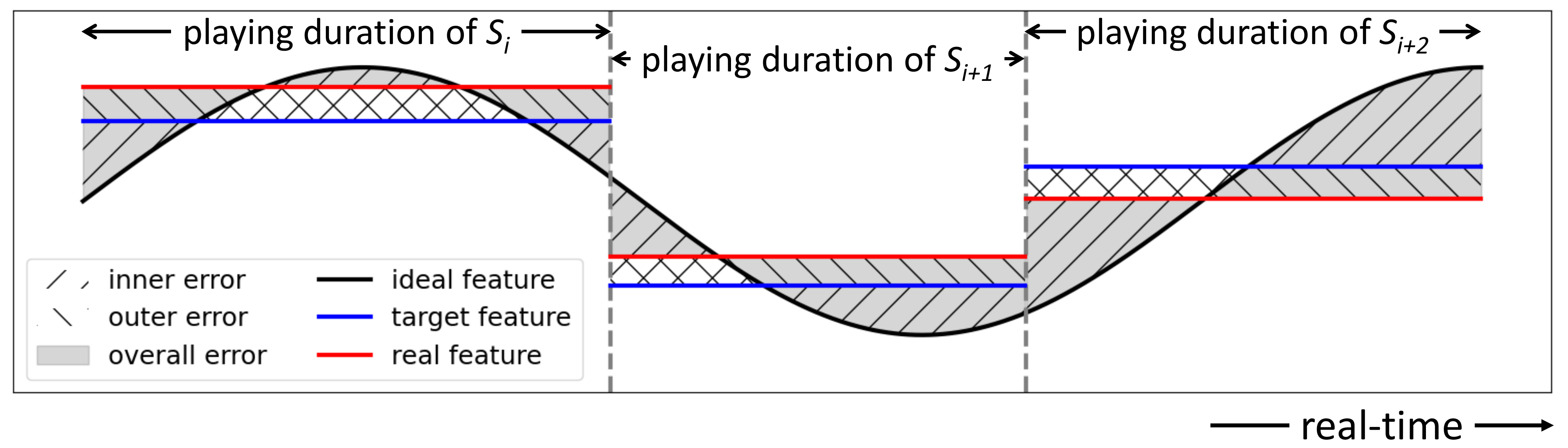}
    \caption{\label{fig:errors}Illustration of the inner error, outer error and overall error formulated in Eqs. \eqref{eq:errin}, \eqref{eq:errout} and \eqref{eq:errall}, respectively.}
\end{figure}

This error $\varepsilon_{inner}$ is called an \textit{inner error} because Eq. \eqref{eq:errin} evaluates the distance between the the targeted feature values and the ideal feature sequence, and cannot be eliminated as $\delta_i$ is always larger than the time unit. In a controller-generator architecture, there can be an additional \textit{outer error} between the targeted feature value and the one of an actually generated segment. 
Hence, the \textit{overall error} does not equal to \textit{inner error}. 
An \textit{outer error}, $\varepsilon_{outer}$, and an \textit{overall error}, $\varepsilon_{all}$, are formulated in Eqs. \eqref{eq:errout} and \eqref{eq:errall}, respectively.
\begin{align}
    \varepsilon_{outer} &= \frac{1}{T(f_{max} - f_{min}) } \sum_{i=1}^N \delta_i |\hat f_i - f_i |. \label{eq:errout} \\
    \varepsilon_{all} &= \frac{1}{T(f_{max} - f_{min}) } \sum_{i=1}^N \left( \sum_{t=b_i}^{b_i+\delta_i} |f_t^* - f_i|\right) \label{eq:errall}.
\end{align}
The notation $f_i$ in Eq \eqref{eq:errout} denotes the real feature value of $S_i$.
As shown in Eqs. \eqref{eq:errin}, \eqref{eq:errout} and \eqref{eq:errall}, those errors are normalised linearly to the range of $[0, 1]$. 
Fig. \ref{fig:errors} illustrates the relationship between those errors.

Although only one level feature is considered as a case study in this work, our formulation can be easily extended to the case of using multiple level features. 

\subsection{Challenges of Music-driven Online Level Feature Control}\label{sec:challenges}
There are at least two challenges of music-driven online level feature control, the uncertainty of play duration and the dilemma of granularity.
The time consumed to play through a segment depends on the player's skill and play style. Therefore, generating level segment from music should be achieved in a player-adaptive way. 
Online level generation usually generates a level segment by segment. Using segments of smaller size is expected to control features more accurately. However, online level generation also requires high generation speed. Using smaller segments will lead to a lower generation speed due to the higher frequency of making control decision. Moreover, extracting features from very small segments does not always make sense. Generating small yet reasonable segments in real-time is not 
trivial. 
To overcome the above challenges, a music-driven online level feature controller is designed and detailed in Section \ref{sec:ol-gen}. 

\section{Online Player-adaptive
Procedural Content Generation via Reinforcement Learning}\label{sec:framework}
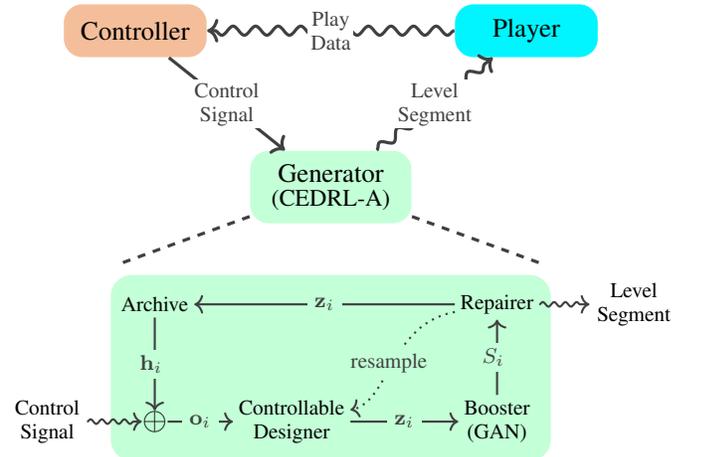
\begin{figure}[htbp]
    \centering
    \definecolor{clr3}{RGB}{195, 255, 215}
\definecolor{clr2}{RGB}{245, 190, 155}
\definecolor{clr1}{RGB}{0, 245, 255}

\begin{tikzpicture}[scale=1.3]
\footnotesize
    \node [rounded corners=7pt, minimum width=54pt, minimum height=20pt, fill=clr1] (P) at (2, 1.6) {\normalsize Player};

    \node [rounded corners=7pt, minimum width=54pt, minimum height=20pt, fill=clr2, align=center] (C) at (-2, 1.6) {\normalsize Controller};

    \node [rounded corners=7pt, inner xsep=8pt, inner ysep=5pt, fill=clr3, align=center] (G) at (0, 0) {\normalsize Generator\\\small(CEDRL-A)};

    \draw [very thick, color=black!75, decorate, decoration={snake, amplitude=.6mm, segment length=3mm, post length=1mm}, ->] (G) -- node[align=center, inner sep=1pt, fill=white]{Level\\ Segment} (P);
    
    \draw [very thick, color=black!75, decorate, decoration={snake, amplitude=.6mm, segment length=3mm, post length=1mm}, ->] (P) -- node[align=center, inner sep=1pt, fill=white]{Play\\ Data} (C);
    
    \draw [very thick, color=black!75, ->] (C) -- node[align=center, inner sep=1pt, fill=white]{Control\\ Signal} (G);

    \draw [very thick, color=black!75, dashed] (G) -- (-2.2, -0.8);
    \draw [very thick, color=black!75, dashed] (G) -- (2.2, -0.8);

    \fill [color=clr3, rounded corners=8pt] (-2.25, -0.9) rectangle (2.25, -2.8);

    \node [align=center, inner sep=2pt] (A) at (-1.8, -1.2) {Archive};
    \node [inner sep=0pt] (cat) at (-1.8, -2.4) {\large $\oplus$};
    \node [align=center, inner sep=2pt] (D) at (-0.4, -2.4) {Controllable \\ Designer};
    \node [inner sep=2pt] (R) at (1.7, -1.2) {Repairer};
    \node [inner sep=2pt, align=center] (B) at (1.7, -2.4) {Booster\\ (GAN)};

    \node [align=center] (S) at (3.1, -1.2) {Level\\ Segment};
    \draw [thick, color=black!75, decorate, decoration={snake, amplitude=.3mm, segment length=1.5mm, post length=.6mm}, ->] (R) -- (S);
    \node [align=center] (CS) at (-2.9, -2.4) {Control\\ Signal};
    \draw [thick, color=black!75, decorate, decoration={snake, amplitude=.3mm, segment length=1.5mm, post length=.6mm}, ->] (CS) -- (cat);
    \draw [thick, color=black!75, ->] (A) -- node[fill=clr3, inner sep=2pt, align=center]{$\mathbf h_i$} (cat);
    \draw [thick, color=black!75, ->] (cat) -- node[fill=clr3, inner sep=0pt, align=center]{$~\mathbf o_i$} (D);
    \draw [thick, color=black!75, ->] (D) -- node[fill=clr3, inner sep=0pt, align=center]{$~\mathbf z_i$} (B);
    \draw [thick, color=black!75, ->] (B) -- node[fill=clr3, inner sep=2pt, align=center]{$S_i$} (R);
    \draw [thick, color=black!75, ->, dotted] (R) ..controls (0.6, -1.4) and (0.6, -2.2).. node[fill=clr3, inner sep=2.3pt, align=center]{resample} (D);
    \draw [thick, color=black!75, ->] (R) -- node[fill=clr3, inner sep=0pt, align=center]{$~\mathbf z_i$} (A);

\end{tikzpicture}
    \caption{Overview of OPARL.
    $\mathbf h_i$, $\mathbf o_i$ and $\mathbf z_i$ refer to the historical latent vectors, observation of designer and latent vector at the $i$th iteration, respectively. \emph{Archive} is a ring queue of latent vectors with capacity $m_G$.}
    \label{fig:overview}
\end{figure}

We propose a framework named online player-adaptive
procedural content generation via reinforcement learning (OPARL) for online level generation from
music. An overview of OPARL is given in Fig. \ref{fig:overview}. OPARL is composed of a feature controller and a segment generator. At each iteration, the controller receives players' play data on the most recent segment and
 determines the control signal, which directs the generator to generate appropriate segments for a specific player. The generator takes a number of previous segments and the control signal as input, and then outputs a new segment. This section explains the framework of OPARL, its implementation and parameter setting used in the experimental studies. More technical details are available in the released project\footref{a}. 

\subsection{Controllable EDRL with Archive}\label{sec:OPARL}

A controllable experience-driven reinforcement learning with archive (CEDRL-A) architecture, extended from the EDRL framework \cite{shu2021experience}, is designed as the generator of OPARL. It models the task of online level generation as a Markov decision process so that a \textit{designer} (known as ``agent'' in \cite{shu2021experience}) can be trained to generate rapidly level segments of high quality. A \textit{booster} (known as ``generator'' in \cite{shu2021experience}) based on GAN is applied to tackle the high-dimensionality of designer's action space.
The booster maps a low-dimensional vector (e.g., latent vector of GAN) to a high-dimensional level segment representation. The observation and action of the designer are the latent vectors of the most recently generated segment and the segment to be generated, respectively. The action space of CEDRL-A is the latent space of the booster. 

Different to EDRL, the designer in CEDRL-A takes additional control signal that describes the desired feature value of the next segment as input, and outputs a segment with a feature value closed to what the input control signal described. This ability of generating segments with a desired feature value is achieved by employing a \textit{controllability} reward while training the generator. Moreover, an \textit{archive} $X_G = \{\mathbf z_j ~|~ j = i-m_G, \cdots, i-1\}$ of $m_G$ previous latent vectors is introduced in CEDRL-A to guarantee the Markov property. Since some reward functions can depend on more than one previous segments, it is necessary to make sure that $m_G$ is larger than or equal to the maximum number of previous segments to be used in any reward function. Let $\mathbf h_i = \mathbf z_{i-m_G} \oplus \cdots \oplus \mathbf z_{i-1}$\footnote{$\oplus$ represents vector concatenating.} represent the concatenated historical latent vectors that a controller can observe and $\mathbf g_i$ represent the vector of the control signal, the controllable designer takes $\mathbf o_i = \mathbf h_i \oplus \mathbf g_i$ as its observation at the $i$th iteration.

Similar to EDRL, CEDRL-A applies a CNet-assisted evolutionary repairer to repair the broken pipes appeared in the generate levels \cite{shu2020novel} and a resampling strategy to ensure the  playability~\cite{shu2021experience}, referred to as \textit{Repairer} in Fig. \ref{fig:overview}. 



\subsection{Training Controllable Designer}

\subsubsection{Reward function}\label{sec:reward}
To make the designer generate segments with feature values closed to the desired ones at each iteration, a \textit{controllability} reward is introduced during  designer training. Let $f(\cdot)$ be the function of computing level feature, the controllability reward is formulated as $$C(S_i) = 1 - \frac{|f(S_i) - \hat f_i|}{f_{max} - f_{min}}.$$

In the implementation of OPARL in this paper,
the difficulty degree of a level segments is considered as the feature of a level segment, quantified as the summation of the number of enemies and the number of empty ground tiles divided by the width of segment, as formulated as
\begin{equation}
    \label{eq:difmetric}
    \mathrm{difficulty} = \frac{\mathrm{\#enemies} + \mathrm{\#empty\_grounds}}{\mathrm{width}}.
\end{equation}
The lower bound and upper bound of difficulty degree are set as $f_{min} = 0$ and $f_{max} = 1$, respectively.

As the targeted feature (difficulty degree in the implementation) directly influences the next segment, the method of sampling features can affect other rewards in some unpredictable way. In some primary attempts, we observe that the difficulty degree usually does not change fast in the online generation phase. Therefore, the feature is sampled uniformly in $[f_{min}, f_{max}]$, and then a Gaussian mutation is applied to the current feature value to get the next feature value recursively. Formally, the feature values sampled during a training epoch are $\hat f_1 \sim U(f_{min}, f_{max})$ and $\hat f_i = \hat f_{i-1} + \xi$ with $\xi \sim N(0, \sigma^2)$, $\forall i>1 $.
In this work, $\sigma$ is arbitrarily set as $0.05$.

\def\nouse{
  \subsubsection{Reward for fun and playability}
  In addition to \textit{controllability}, reward functions  of \textit{fun} and \textit{playability} proposed in \cite{shu2021experience} are also employed to ensure player experience. The reward for \textit{fun} guides the designer to generate segments neither too diverge nor too similar to a part of the newly generated segment.
  The reward function \textit{fun}~\cite{shu2020novel} is defined as:
  \begin{equation}
    F(S_i) = \begin{cases}
      -[D(S_i) - u]^2, & \text{if } D(S_i) > u, \\
      -[D(S_i) - l]^2, & \text{if } D(S_i) < l, \\
      0, & \text{otherwise,}
    \end{cases}
  \end{equation}

  where $l=0.26$ and $u=0.94$ are hyper-parameters that describe the lower bound and upper bound of acceptable novelty score of the newly generated segment, respectively. $D(\cdot)$ is a metric for novelty~\cite{shu2020novel}, formulated as:
  \begin{equation}
      \label{eq:diversity}
      D(S_i) = \frac{1}{n} \sum_{k=1}^{n} \TPKL(S_i, H_i[k s]).
  \end{equation}
  $H_i$ denotes the historical records of concatenated previous segments when the $i$th segment is being generated. $H_i[x]$ denotes the segment that is $x$ tiles far from the end of $H_i$ (i.e., $H_i[0] = S_{i-1}$). $\TPKL(\cdot, \cdot)$ denotes a $2 \times 2$ tile-pattern KL-divergence measured as in \cite{lucas2019tile,shu2021experience}.
  The number of segments to be compared $n$ is set as $3$. $s$ is the stride of sliding windows for selecting segments and set as the half of the segment width. All the hyper-parameters used in calculating \textit{fun} are kept the same as in \cite{shu2021experience}, expect that the segment width is $28$ here.

  The novelty metric defined in Eq. \eqref{eq:diversity} is slight different from the one defined in \cite{shu2021experience}. Eq. \eqref{eq:diversity} computes the dissimilarity starting from $S_{i-1}$ instead of $S_i$ itself, because results of primary experiments show that computing tile-pattern divergence with segments the same or overlapped with $S_i$ will always result in small values and does not improve the content quality.
}

We also employ \textit{fun} and \textit{playability} introduced in the work of \cite{shu2021experience} as additional reward terms so that the generated levels are fun and playable. 
\def\tooDetailed{
For \textit{fun}, slightly different from the original, we start the sliding window from $S_{i-1}$ rather than $S_i$ to avoid computing tile-pattern KL-divergence between overlapped segments, because we find in such case tile-pattern KL-divergence always produce very small value and does not always make sense. When determining \textit{playability}, different from \cite{shu2021experience}, we do not terminate the MDP once an unplayable segment is generated. Otherwise, the designer is encouraged to maximise the playable length of a generated level. However, in online generation phase, the resample strategy can be used to ensure the playability. It is better to minimise the times of resampling (i.e., the number of unplayable segments) rather than maximising the playable length.} 
The playability of segment $S_i$ is checked by simulating $S_{i-1} + S_i$ with an $A^*$ agent that won the 2009 Mario AI Competition \cite{togelius20102009}. In our work, the \textit{playability} is set as $0$ if the newly generated segment is playable, otherwise $-1$.

\subsubsection{Training Designer with Soft Actor-critic}
Among the reward terms, \textit{controllability} does not depend on previous segments, and \textit{playability} depends on no more than one previous segment. However, in our setting, computing \textit{fun} requires $2$ previous segments ($S_{i-1}$ and $S_i$). Thus, the capacity of the generator's archive $m_G$ is $2$. The designer uses a multi-layer perception (MLP) model. The booster uses a latent space of $\mathbb R^{20}$.
The targeted difficulty degree is duplicated by $12$ times to increase the number of connections in the MLP on the inputted targeted difficulty.
As a result, the control signal $\mathbf g_i$ has a dimensionality of $12$ and the designer's observation $\mathbf o_i$ is a $52$-dimensional vector. The action space is $[-1, 1]^{20}$. 

We implement a parallel-environment version of soft actor-critic (SAC)~\cite{haarnoja2018soft,haarnoja2018sac} as an OpenAI Gym interface~\cite{brockman2016openai} with $50$ synchronous sub-environments to train the designer for 1 million time steps in total. The actor and critics use MLP model with 3 hidden layers of $256$ neurons. Our implementation of SAC updates the models $10$ times using randomly sampled batches of size $384$ every time $100$ transitions are collected. The automating entropy adjustment~\cite{haarnoja2018sac} is used with a targeted entropy $\bar{\mathcal H} = -20$, as recommended in \cite{haarnoja2018sac}. 
The smoothing coefficient $\tau$ is set as $0.005$. The discounted factor $\gamma$ greatly affects the performance of reinforcement learning algorithms. Empirically, a smaller $\gamma$ (comparing with $\gamma = 0.99$ in many baselines) is better. In all the experiments reported in this paper, $\gamma= 0.7$. Each epoch terminates after $50$ segments are generated. For the last transition, the target Q-value is computed as $r + \gamma Q(s', \tilde a')$ rather than $r$ since the training aims at endless online generation.

\subsection{Online Player-Adaptive Level Generation with Controller \label{sec:ol-gen}}

Given a piece of music, our music-driven feature controller will first compute the ideal feature sequence $\mathcal F^*$ during initialisation, and then store it. In addition to $\mathcal F^*$, the control policy keeps an archive $X_C = \{\langle \hat f_j, \delta_j \rangle ~|~ j = i-m_C-1, \cdots, i-2\}$ to keep at most $m_C$ entries of the previous segments. Each entry is composed of the targeted feature value and play duration of a segment. The last entry of $X_C$ is about the $(i-2)$th segment at the $i$th iteration because $\delta_{i-1}$ is unknown when determining $\hat f_i$.

\begin{algorithm}[htbp]
\caption{LS-KNN\label{algo:rtfc-knn}. In the experiments, $\#\mathrm{trial} = 50$, $k = 5$ and $\sigma_c = 0.02$.}
\begin{spacing}{1.08}
\begin{algorithmic}[1]
    \Require $\hat f_{i-1}$, $b_{i-1}$, $\mathcal F^*$, $X_C$  {\Comment{\textit{Inputs}}}
    \Require $\mathrm{\#trials}$, $k$, $\sigma_c$  {\Comment{\textit{Hyper-parameters}}}

    \Ensure $\hat f_{i}$
    
    \State $\tilde b_i \gets b_{i-1} + \Est(\hat f_{i-1}; X_C)$ {\Comment{\textit{Estimate beginning time}}}

    \State $\hat f_i \gets \hat f_{i-1}$
    \State $\tilde \delta_i \gets \Est(\hat f_i; X_C)$
    \State $\tilde \varepsilon \gets \Eval(\hat f_i; \tilde b_i, \tilde \delta_i, \mathcal F^*)$ {\Comment{\textit{Least estimated error found}}}
    \Repeat
        \State $\hat f_i' \gets \hat f_i + \xi, ~~~ \xi \sim N(0, \sigma_c^2)$ {\Comment{\textit{Do local search}}}
        \State $\tilde \delta_i \gets \Est(\hat f_i; X_C)$ 
        \State $\tilde \varepsilon' \gets \Eval(\hat f_i; \tilde b_i, \tilde \delta_i, \mathcal F^*)$
        \If {$\tilde \varepsilon' < \tilde \varepsilon $}
            \State $\hat f_i \gets \hat f_i'$
            \State $\tilde \varepsilon \gets \tilde \varepsilon'$
        \EndIf
    \Until {has looped for $\mathrm{\#trials}$ times}
    \State \Return $\hat f_i$
\end{algorithmic}
\end{spacing}
\end{algorithm}

A simple algorithm named local search with KNN-based estimation (LS-KNN, Algorithm \ref{algo:rtfc-knn}) is designed as the online control policy for determining targeted feature values. When determining a targeted feature value for the $i$th segment, LS-KNN executes local search for $\mathrm{\#trails}$ generations starting from the last one $\hat f_{i-1}$, and picks up the best value of $\hat f_i$ found according to the estimated individual \textit{inner error} using KNN-based prediction of play duration for the $(i-1)$th and $i$th segments. In our case, an individual is mutated by adding a Gaussian noise with a standard deviation of $\sigma_c$. The play duration is estimated with:
\begin{equation*}
    \label{eq:est}
    \Est(\hat f; X_C) = \frac{1}{k} \sum\nolimits_{j \in J} \delta_j,
\end{equation*}
where $J$ is the set of $k$ nearest neighbours in terms of $|\hat f_j - \hat f|$ within $X_C$.
With $\Est(\hat f; X_C)$, we can further estimate the time that the $i$th segment $\tilde b_i$ starts to be played. With $\tilde \delta_i$, the estimated play duration respected to $\hat f_i$, a feature can be evaluated using: 
\begin{equation*}
    \Eval(\hat f_i; \tilde b_i, \tilde \delta_i, \mathcal F^*) = \frac{1}{\tilde \delta_i} \sum\nolimits_{t=\tilde b}^{\tilde b_i + \tilde \delta_i} |f^*_t - \hat f_i|.
\end{equation*}


Algorithm \ref{algo:rtfc-knn} details the implementation of the control policy. LS-KNN is proposed with two assumptions: (i) an ideal feature sequence won't change fast, thus the ``optimal'' $\hat f_i$ should not be far from $\hat f_{i-1}$; (ii) the difficulty implies the play duration, which is true for many platformer games.
The former is somehow generic in online level generation because fast changes of features may be harmful to the coherence of levels. It is the reason of starting the local search from $\hat f_{i-1}$ in Algorithm \ref{algo:rtfc-knn}. The latter motivates the estimation of play duration according to the records organised by $\langle \hat f_j, \delta_j \rangle$.


LS-KNN is easy to implement and can always be used directly without training or other preparation. Furthermore, though not explicit, LS-KNN is well player-adaptive since the KNN-based estimation is applied based on the specific player's play data collected online. Our implementation with LS-KNN policy achieves significant performance in the simulation-based experiments detailed in Section \ref{sec:exp}.


\subsection{Implementation of Booster}
Our booster is a variation of MarioGAN~\cite{volz2018evolving}. Differ from the original neural network architecture, our GAN model uses fractional-convolutional layers~\cite{radford2015unsupervised} with kernel size, stride and padding of $\langle (4, 4), (1, 1), 0 \rangle$, $\langle (3, 3), (2, 2), 1 \rangle$, $\langle (4, 4), (2, 2), 1 \rangle$ and $\langle (3, 4), (1, 2), 1 \rangle$, respectively, to directly obtain an output of size $14 \times 28$ without clipping.
Besides, latent vectors of length $20$ are used. The generator  and discriminator are trained for $5$ times and $1$ time at each iteration, respectively, on the $13$ human-designed levels without bullet bills
provided in the Video Game Level Corpus~\cite{summerville2016vglc}.

\section{Experimental Study\label{sec:exp}}
To evaluate the effectiveness of our approach and implemented algorithms in optimising different objectives, the designer is trained with all the possible combinations of the three reward terms presented in Section \ref{sec:reward} with the same weights, and evaluated with training environments and online generation environments, respectively. 

\begin{figure}[t]
  \centering
  \includegraphics[width=\linewidth]{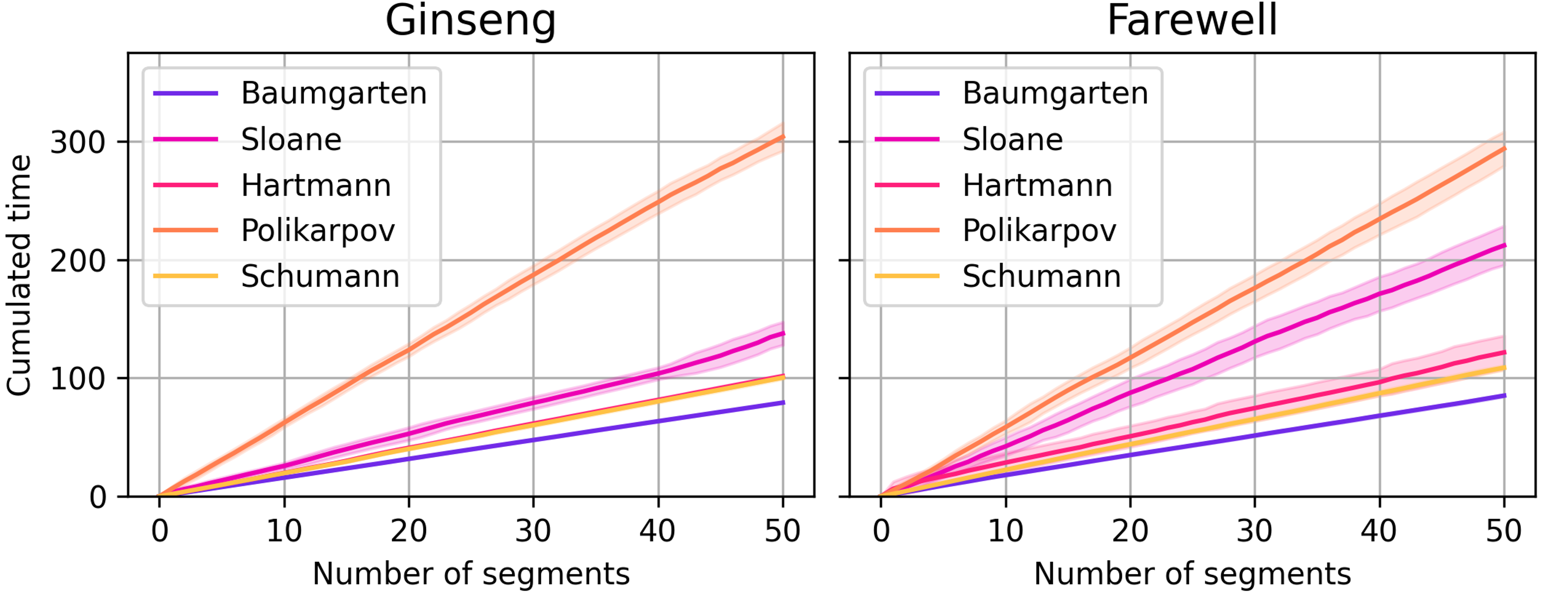}
  \caption{Cumulative play duration of each agent on levels generated from the same initial segment, averaged over 30 independent trails. Shadows indicate the standard deviation.}
  \label{fig:time}
\end{figure}

To test the robustness of our method, five different agents in the 2009 Mario AI Competition~\cite{togelius20102009}, namely Baumgarten's (the aforementioned $A^*$ agent), Sloane's, Hartmann's, Polikarpov's and Schumann's agents, are used as the simulated player, and two different pieces of music, \textit{Ginseng}\footnote{From the original sound track of commercial platformer game \textit{Electronic Super Joy: Groove City} (Michael Todd, 2014).} (EnV, 2014) and \textit{Farewell}\footnote{From the original sound track of commercial platformer game \textit{Celeste} (Matt Makes Games Inc., 2018).} (Raine, 2019), are used. Fig. \ref{fig:time}. shows the five agents' play duration on each segments of an level generated online by OPARL using a designer trained with the summation of \textit{controllability}, \textit{fun} and \textit{playability}. Those agents actually play levels with different speeds.
Our experiments are simulated based on the Mario-AI-Framework\footnote{\url{https://github.com/amidos2006/Mario-AI-Framework}}.


\begin{figure}[htbp]
    \centering
    \includegraphics[width=\linewidth, trim=90 0 108 0, clip]{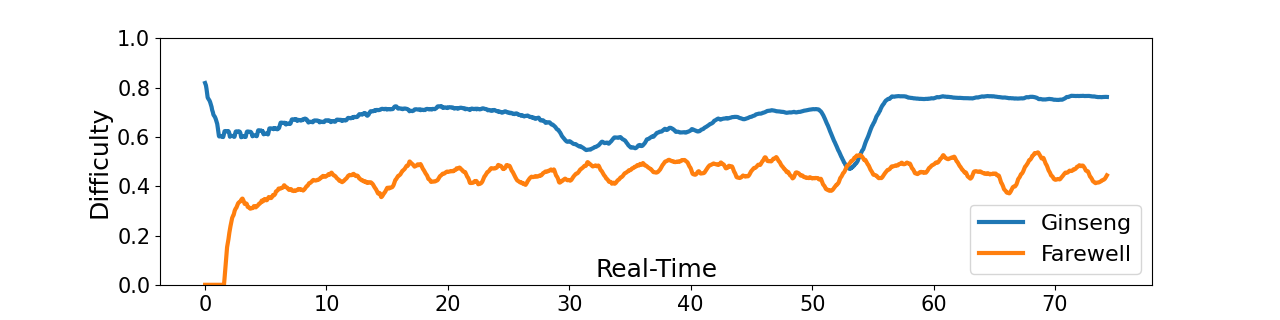}
    \caption{Parts of converted ideal difficulty sequences extracted from the music used in our experiments.}
    \label{fig:idc}
\end{figure}

Root-mean-square energy features of music are extracted throughout time and mapped to difficulty degrees as follows. First, the degrees are re-scaled by taking logarithm based on $10$, then clipped within $[-2.5, 0]$ and mapped into $[0, 1]$ linearly, finally, smoothed through computing mean values in an $1$-stride sliding window of size $100$. The music feature extraction is done through \textit{Librosa} library~\cite{mcfee2015librosa}, with a default time unit of $0.02322$s. Parts of the resulted ideal difficulty sequences are illustrated in Fig. \ref{fig:idc}. Demos of levels generated from different music are available in the released project\footref{a}.

\subsection{Evaluation of Effectiveness}

The trained designers are evaluated with three metrics, $\sqrt{-F}$, the mean distance of $D(S_i)$ out of the range $[l, u]$, $-P$, the negative number of \textit{playability} to indicate the ratio of unplayable segments, and $1-C$ to indicate the mean error between level feature and targeted feature value in training environment.
\def\removedforspace{
\begin{itemize}
    \item $\sqrt{-F}$: The mean distance of $D(S_i)$ out of the range $[l, u]$.
    \item $-P$: Opposite number of \textit{playability} reward. Indicate the ratio of unplayable segments.
    \item $1-C$: Indicate the mean error between level feature and targeted feature value in training environment.
\end{itemize}}
Those metrics are averaged over all iterations (time steps) and normalised in a similar scale.

To further evaluate the performance of online generation, the designer is tested using Baumgarten's agent as simulated player and \textit{Ginseng} as input music. The values of $\sqrt{-F}$, $-P$,
$\varepsilon_{inner}$, $\varepsilon_{outer}$, $\varepsilon_{all}$, and an additional metric $Div$, which is used to measure the \textit{diversity} of levels generated in different runs, are reported in Table \ref{tab:train}. $Div$ is computed as the mean ratio of tiles that are different in pairs of levels generated in different runs. The values of $\sqrt{-F}$ and $-P$ in the online evaluation may be different to the ones in the offline evaluation due to resampling and different targeted feature values.


\renewcommand\arraystretch{1.15}
\begin{table*}[htbp]
    \centering
    \caption{Evaluation of designers trained with different reward functions. All values are averaged over $100$ independent trials. $Div$ is to be maximised, all the other metrics are to be minimised with strict lower bound of $0$. Cells filled with $-/-$ are meaningless. The best and worst results are highlighted with bold and italic, respectively.}
    \label{tab:train}
    \small
    \setlength\tabcolsep{4.5pt}    
    \begin{tabular}{c|ccc|cccccc}
        \toprule[1.2pt]
        \multirow{2}{*}{\textbf{Designer}} 
        & \multicolumn{3}{c|}{\textbf{Training Environment}} & \multicolumn{6}{c}{\textbf{Online Generation}} \\
        & $\sqrt{-F}$\scriptsize{$(10^{-2})$} & $-P$\scriptsize{$(10^{-2})$} & $1-C$\scriptsize{$(10^{-2})$} & $\sqrt{-F}$\scriptsize{$(10^{-2})$} & $-P$\scriptsize{$(10^{-2})$} & $\varepsilon_{inner}$\scriptsize{$(10^{-2})$} & $\varepsilon_{outer}$\scriptsize{$(10^{-2})$} & $\varepsilon_{all}$\scriptsize{$(10^{-2})$} & $Div$\\
        \midrule[0.8pt]
        $F$ & \textbf{1.08 $\pm$ 0.34} & 9.12 $\pm$ 4.03 & $-/-$ & \textbf{1.06 $\pm$ 0.35} & 9.06 $\pm$ 4.22 & $-/-$ & $-/-$ & 43.6 $\pm$ 1.76 & \textbf{0.066} \\
        $P$ & 11.2 $\pm$ 1.81 & \textbf{0.22 $\pm$ 0.63} & $-/-$ & 11.2 $\pm$ 1.90 & \textbf{0.00 $\pm$ 0.00} & $-/-$ & $-/-$ & 48.2 $\pm$ 1.01 & \textbf{0.066} \\
        $F{+}P$ & 2.06 $\pm$ 0.58 & 0.42 $\pm$ 0.82 & $-/-$ & 2.29 $\pm$ 0.58 & \textbf{0.00 $\pm$ 0.00} & $-/-$ & $-/-$ & \textit{49.9 $\pm$ 0.92} & 0.046 \\
        \midrule[0.8pt]
        $C$ & \textit{16.8 $\pm$ 3.27} & \textit{46.7 $\pm$ 21.2} & \textbf{1.90 $\pm$ 0.23} & \textit{19.2 $\pm$ 0.99} & \textit{29.4 $\pm$ 5.49} & \textit{1.16 $\pm$ 0.10} & \textbf{1.59 $\pm$ 0.22} & \textbf{2.24 $\pm$ 0.19} & 0.056 \\
        $C{+}F$ & 4.48 $\pm$ 2.13 & 15.3 $\pm$ 6.76 & 2.88 $\pm$ 0.49 & 1.83 $\pm$ 0.52 & 5.66 $\pm$ 3.12 & 0.89 $\pm$ 0.04 & 2.69 $\pm$ 0.26 & 2.95 $\pm$ 0.24 & 0.057 \\
        $C{+}P$ & 11.9 $\pm$ 3.41 & 0.40 $\pm$ 0.85 & 3.77 $\pm$ 0.70 & 17.5 $\pm$ 1.59 & 0.44 $\pm$ 0.96 & 0.89 $\pm$ 0.04 & \textit{4.82 $\pm$ 0.73} & 4.99 $\pm$ 0.80 & 0.062 \\
        $C{+}F{+}P$ & 6.81 $\pm$ 1.31 & 0.40 $\pm$ 0.94 & \textit{4.36 $\pm$ 0.56} & 7.86 $\pm$ 0.97 & 0.02 $\pm$ 0.20 & \textbf{0.87 $\pm$ 0.03} & 4.74 $\pm$ 0.54 & 4.93 $\pm$ 0.51 & \textit{0.035} \\
        \bottomrule[1pt]
    \end{tabular}
\end{table*}

Table \ref{tab:train} shows the experimental results.
The designers trained with \textit{controllability} generally achieve very low \textit{overall error} in the online generation tests. The main source of \textit{overall error} is the \textit{outer error}, i.e., the error between the targeted feature value produced by controller and the feature of actually generated segment. The value of $1-C$ closed to $\varepsilon_{outer}$ means that our method of sampling  targeted features is effective. The designer trained with only \textit{fun} reward achieves a great score on $\sqrt{-F}$. However, when \textit{controllability} is employed, the score of \textit{fun} deteriorates a lot. This phenomenon indicates that the objective of \textit{fun} and \textit{controllability} conflict. 
Moreover, the \textit{controllability} deteriorates less comparing with the designer trained with \textit{controllability} only. It is probably because the reward of \textit{fun} uses a quadratic form while \textit{controllability} uses a linear form. That means designer finds it better to optimise \textit{controllability} to get a higher summation of reward terms.

All the designers trained with \textit{playability} well ensure the playabiltiy of generated levels, while designers generally assure better playability with the help of resampling. A merit attention finding is that the designer trained with $F$ and the designer trained with $C+F+P$ do not get notable better $P$ value in the online generation tests. A possible reason is that those designers lack of randomness when taking actions.
That means if they generate an unplayable segment, no matter how many times the re-generation is executed, they will still generate unplayable segments. A future work is finding out why the phenomenon only appears on those two designers.


As a conclusion, our implemented framework optimises the reward functions effectively. The designer trained with $C+F+P$ balances different objectives and can be a good choice for online level generation from music.

\subsection{Evaluation of Robustness}
\begin{figure}[htbp]
  \centering
  \subfigure[Evaluation scores of $1 - 
  \varepsilon_{all}$ for different simulated players\label{fig:bar-err}.] {\includegraphics[width=\linewidth]{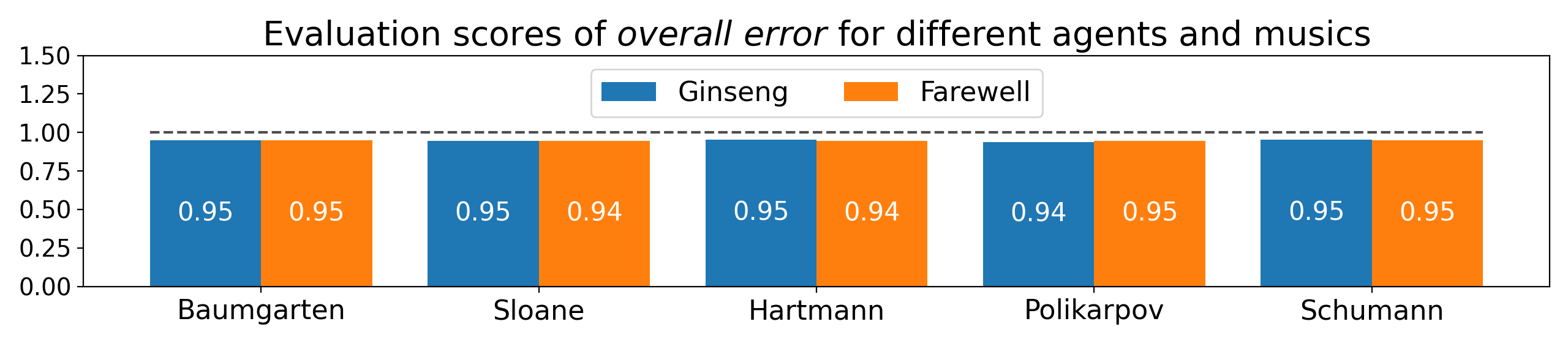}} \\
  \subfigure[Evaluation scores of $1 - \sqrt{-F}$ for different simulated players.\label{fig:bar-fun}] {\includegraphics[width=\linewidth]{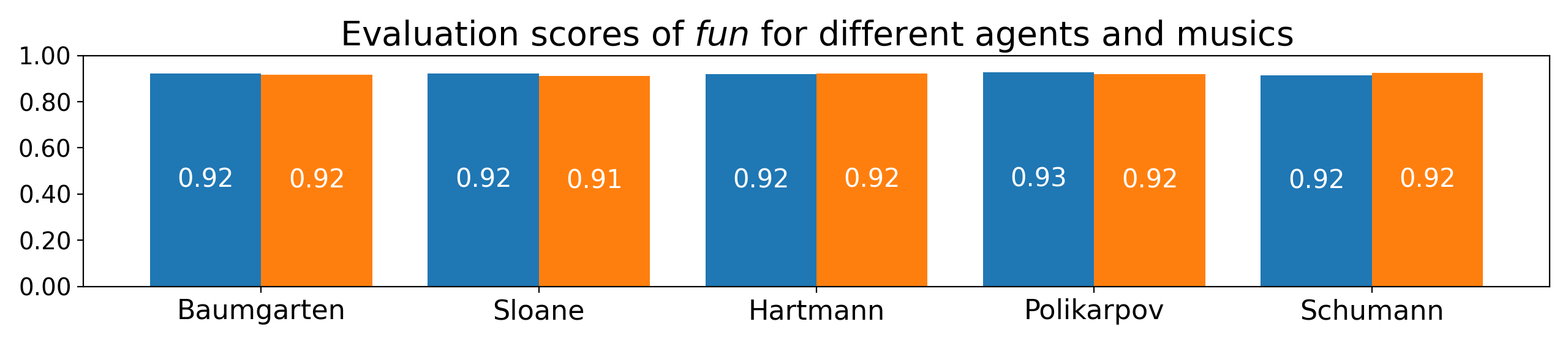}} \\
  \subfigure[Evaluation scores of $Div$ for different simulated players.\label{fig:bar-div}] {\includegraphics[width=\linewidth]{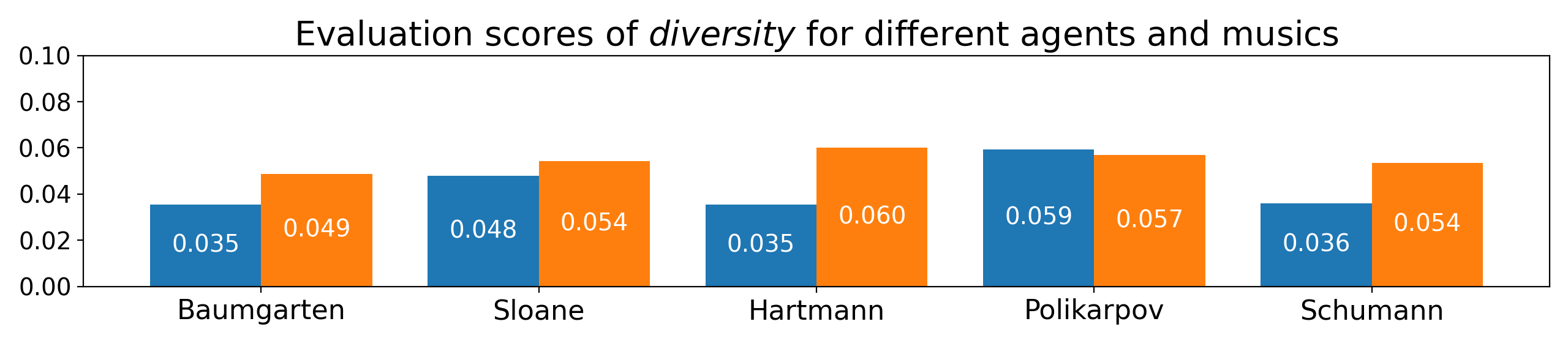}} 
  \caption{Values of \textit{fun}, \textit{controllability}, \textit{diversity} evaluated on designers trained with different agents as player and different musics. Each value is averaged over 30 independent trials.}
  \label{fig:bars}
\end{figure}

Fig. \ref{fig:bars} plots the \textit{overall error}, \textit{fun} and \textit{diversity} evaluated on the designer trained with $C+F+P$ as reward for the five agents and two different pieces of music. The \textit{overall error} and \textit{fun} are plotted as $1 - \epsilon_{all}$ and $1 - \sqrt{-F}$ for better intelligibility. 
According to Figs. \ref{fig:bar-err} and \ref{fig:bar-fun}, our method achieves very similar and high performances of \textit{overall error} and \textit{fun}, and is robust for players with different play speed. The diversity scores of levels generated for different players and musics vary significantly. According to Figs. \ref{fig:time} and \ref{fig:idc}, the diversity of levels generated by OPARL may be positively correlated with the fluctuation degree of music and the variance of play duration.

\begin{figure*}[htbp]
    \centering
    \subfigure[Level Generated with \textit{Ginseng} for Hartmann's agent ]{\includegraphics[width=0.48\linewidth]{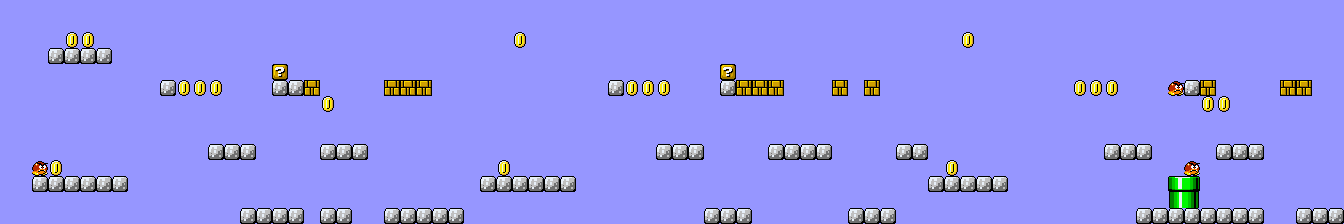}}  \hspace{7pt}
    \subfigure[Level Generated with \textit{Farewell} for Hartmann's agent ]{\includegraphics[width=0.48\linewidth]{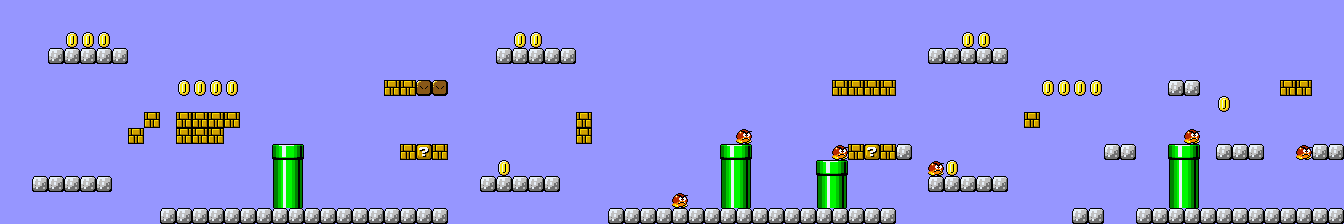}} \\
    \subfigure[Level Generated with \textit{Ginseng} for Sloane's agent]{\includegraphics[width=0.48\linewidth]{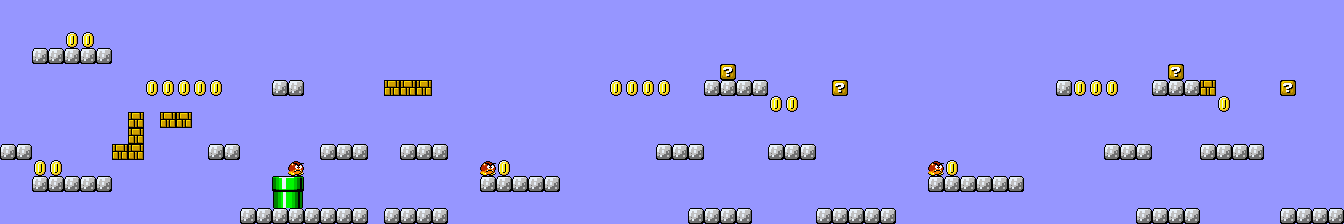}}  \hspace{3pt}
    \subfigure[Level Generated with \textit{Farewell} for Sloane's agent]{\includegraphics[width=0.48\linewidth]{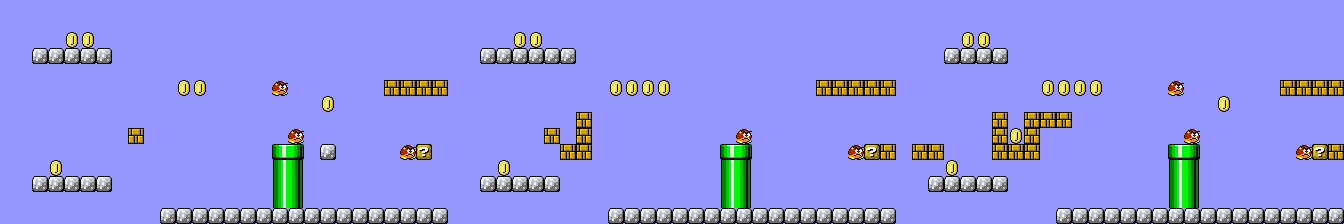}} \\

    \caption{Levels generated from an identical initial segment using the designer trained with $F{+}C{+}P$ as reward, different musics and different agents as player. Levels are captured from the $21$th segments, as the controller's archive is fulfilled after $20$ segments were generated.}
    \label{fig:lvl-imgs}
\end{figure*}

Fig. \ref{fig:lvl-imgs} presents segments captured from the levels generated with the same $S_0$ (i.e., initial segment) for different agents and different music. 
Fig. \ref{fig:illustration} uses an illustration to explain how OPARL generates different levels for different players. It is shown in Fig. \ref{fig:lvl-imgs} that levels generated from \textit{Ginseng} are generally harder than the ones generated from \textit{Farewell}, as the ideal difficulty sequence derived from \textit{Ginseng} is generally larger than the one from \textit{Farewell} (cf. Fig. \ref{fig:idc}). The levels generated for different agents with the same music are similar. It may be explained by using the same starting segment to generate those levels. 
To summarise, our method can adapt well different players and is robust to different music.

\section{Conclusion}\label{sec:conclusion}
This paper formulates the problem of online level generation from music, and proposes an online player-adaptive procedural content generation via reinforcement learning (OPARL) framework composed of a novel CEDRL-A generator and a novel LS-KNN controller to achieve online level generation from music. Experimental results show that the implementation of OPARL can generate in real-time SMB levels with segment-wise features closed to an ideal difficulty sequence derived from a piece of music. The resulted generation system can also guarantee the playability.
The training algorithm implemented in this paper achieves considerable performance and can be used as a baseline in further studies. Moreover, our framework is flexible since the controller and the generator are decoupled. The CEDRL-A generator in our framework can be integrated with other controllers like DDA controller for different aspects of player-adaptation.

In this paper, our proposed approaches are verified with simulation-based studies. One of the future work is conducting human tests. 
As another future work, new ways of mapping multiple features of both levels and music can be studied for the purpose of achieving better consistence between play experience and music.

\section*{Acknowledgement}
The authors would like to thank the anonymous reviewers for their valuable comments.

\balance
\bibliographystyle{IEEEtran}
\bibliography{main}

\end{document}